# Read classification using semi-supervised deep learning

Tomislav Šebrek[1], Jan Tomljanović[1], Josip Krapac[1], Mile Šikić[1,2]

[1]University of Zagreb Faculty of Electrical Engineering and Computing, Zagreb, Croatia
[2]Bioinformatics Institute, A*STAR, Singapore
```
tomislavsebrek@gmail.com
    jan.tomlj@gmail.com
    josip.krapac@fer.hr
    mile.sikic@fer.hr
```

**Abstract.** In this paper, we propose a semi-supervised deep learning method for detecting the specific types of reads that impede the *de novo* genome assembly process. Instead of dealing directly with sequenced reads, we analyze their coverage graphs converted to 1D-signals. We noticed that specific signal patterns occur in each relevant class of reads. Semi-supervised approach is chosen because manually labelling the data is a very slow and tedious process, so our goal was to facilitate the assembly process with as little labeled data as possible. We tested two models to learn patterns in the coverage graphs: M1+M2 and semi-GAN. We evaluated the performance of each model based on a manually labeled dataset that comprises various reads from multiple reference genomes with respect to the number of labeled examples that were used during the training process. In addition, we embedded our detection in the assembly process which improved the quality of assemblies.

**Keywords:** Deep-learning · Semi-supervised learning · *De novo* assembly · Chimeric read · Repeat read

## 1 Introduction

*De novo* genome assembly is a process of creating a genome sequence out of short sequenced fragments called reads [1]. Computer algorithms assemble genomes by putting overlapping reads into their correct order. They aim to provide complete, accurate and contiguous assembly. However, some type of reads, that will be referred to as chimeric and repeat reads, prevent the successful assembly [2]. Chimeric reads represent fused sequences from two or even more distinct portions of the genome, while repeat reads contain repetitive regions of the original genome. Reads that are neither chimeric nor repeats will be referred to, in this work, as regular reads.

In this work, we will focus on long reads only. We have analyzed reads of the well-known *E. coli* genome produced by both PacBio and ONT technologies and noticed that unique characteristics of problematic reads are reflected in their coverage graphs. A coverage graph is a graph which is created for each read by counting the number of times each nucleotide pair in the read overlaps with some other read. Examples of coverage graphs can be found in Fig. 1.

We argue that it is possible to classify reads using their coverage graphs as 1D-signals and we will approach this problem with generative deep learning models. When manually classifying reads, most of the reads can be labeled with high confidence using only the features from the coverage graph, but the final decision entails using information from mapping the read to the reference genome. Since this process is quite laborious we will propose a method that is based on a semi-supervised approach with as little labeled data as possible. Furthermore, this approach could be used on reads where the reference genome is unknown and is yet to be assembled.

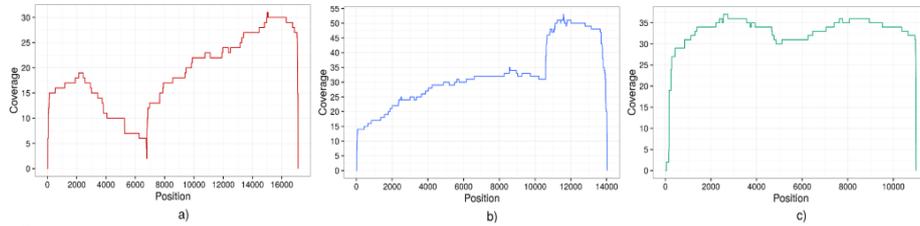

**Fig. 1.** Coverage graph of a) chimeric read, b) repeat read and c) regular read. It can be noticed that chimeric reads are characterized by the sudden drop, followed by the sudden raise of the coverage. Furthermore, repeat read is characterized by a greater coverage on the one side of a coverage graph (example of a right-repeat is shown). Finally, coverage graph of a regular read does not contain any prominent aberrations.

Finally, we argue that omitting chimeric reads and carefully handling repeat reads will lead to the improvement in the assembly process. Since chimeric reads are not part of the reference genome they can cause junction of disjointed regions of the reference genome. Therefore, it is justified to omit them before processing overlaps among reads. On the other hand, repeat reads contain correct and pertinent information about the reference genome, so they need to be considered in order to assemble the genome. However, we assume that omitting all overlaps between left-repeat and right-repeat reads will improve the assembly process.

## 2    Methodology

We tested the following two semi-supervised deep learning models: M1+M2 model and semi-supervised generative adversarial network.

### 2.1    M1+M2

M1+M2 model is a combination of a variational autoencoder which is used as a feature extractor and is referred to as M1 and a generative semi-supervised model which is referred to as M2 [3]. Both M1 and M2 models are based on an approximate inference so they can be described through generative model (P model) and the variational approximation of the intractable posterior (Q model).

P model of M1 is defined as $p(z) \cdot p_\theta(x|z)$, and the Q model is defined as $q_\varphi(z|x)$. It can be noticed that conditional distributions are parameterized with $\theta$ and $\varphi$ and represent encoder and decoder which are implemented as deep neural networks.

Model M2 includes the labels in the generative model. Therefore, P model is defined as $p(z) \cdot p(y) \cdot p_\theta(x|z, y)$ and the Q model is defined as $q(z, y|x) = q(y|x) \cdot q(z|x, y)$. Conditional distribution of P model is referred to as the encoder and the conditional distributions of Q model are referred to as the classification part and the decoder.

### 2.2 Semi-supervised generative adversarial networks (semi-GAN)

The idea of GANs is to have two players, the generator G(z) and the discriminator D(x), that are playing a game against one another [4]. G takes latent vector as input and produces the D-dimensional output which represents generated example. D takes real or generated example and outputs a single scalar that represents the probability that the input example is real. We train the discriminator to maximize the probability of distinguishing between the real and generated dataset and the generator to do the opposite. This can be seen as a min-max game played with a value function V, shown in formula 1.

$$V(D, G) = E_{x \sim p_{data}(x)}[\log D(x)] + E_{z \sim p_z(z)}[1 - \log D(G(z))] \quad (1)$$

In order to support semi-supervised learning, we need to make some changes to the adversarial model. The discriminator will produce K+1 outputs in a form $[p_{y_1}, p_{y_2}, \ldots, p_{y_K}, p_{fake}]$, where K is the number of classes. $p_{y_i}$ represents the probability that an example is real and belongs to the i-th class, and $p_{fake}$ represents the probability that an example is fake [5]. The generator is trained to minimize $p_{fake}$ for generated examples, and the discriminator is trained to maximize $p_{y_k}$ where $y_k$ is the corresponding class.

### 2.3 Dataset

We have not used raw coverage graphs as inputs for any model, but rather the signals that were the result of applying the following chain of operations to the coverage graphs: normalization and down-sampling to a fixed length.

All models were trained on the dataset that was composed of signals obtained from multiple reference genomes. GraphMap tool [6] was used to generate overlaps of reads.

Since all the classes are not equally frequent, we have not used all the signals. The repeat reads are not as recurrent as the regular ones, and the chimeric reads are somewhat obscure, so running a model on the entire dataset would have an unavoidable bias towards the regular reads. To solve this problem and balance the dataset we have de-signed a heuristic for detecting the class label of a signal based on the following features: the incline of certain parts of the coverage graph, the mean value of the left and right side of the coverage graph etc. The heuristic results are unreliable but useful.

The somewhat balanced dataset consisted of 21600 unlabeled and 500 manually labeled examples where labeled examples were divided into the training set (260) and the test set (240).

## 3  Results

Signals were classified into four classes: chimeric, left-repeat, right-repeat and regular. We have compared the performance of both semi-supervised approaches with the respect to the number of labeled data samples that were used during the training process. Results calculated using the test set are presented in Table 1. Results obtained by the semi-supervised approaches are compared to a simple feed-forward convolutional neural network which only used labeled data for training.

**Table 1.** F-score for the feed-forward convolutional neural network (FF), and semi-supervised models M1+M2 and semi-GAN with respect to the number of labeled data samples used during the training process (N)

| N  | FF     | M1+M2  | semi-GAN |
|----|--------|--------|----------|
| 15 | 0.5833 | 0.8750 | 0.9375   |
| 30 | 0.7000 | 0.9083 | 0.9699   |
| 70 | 0.7408 | 0.9341 | 0.9708   |

Latent vectors produced by the models are visualized using the t-SNE algorithm [7] which maps high-dimensional vectors to two- or three-dimensional space. Fig. 2. and Fig. 3. show the results.

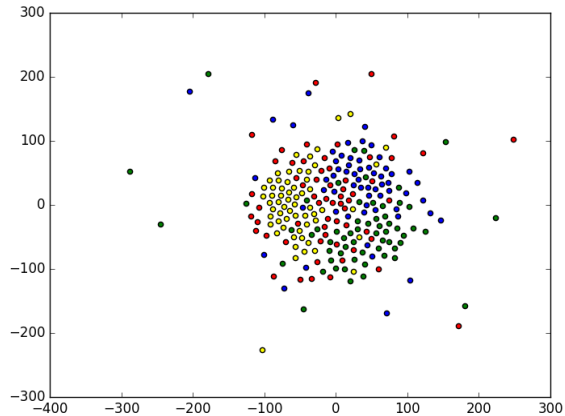

**Fig. 2.** Visualization of the latent vectors of size 10 using t-SNE extracted by the M1+M2. The red, green, blue and yellow circles represent examples from the dataset that were labeled as chimeric reads, left-repeats, right-repeats and regular reads, respectively.

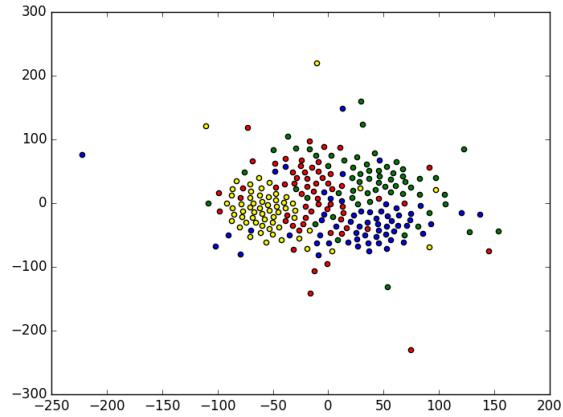

**Fig. 3.** Visualization of the latent vectors of size 10 using t-SNE extracted by the semi-GAN. The red, green, blue and yellow circles represent examples from the dataset that were labeled as chimeric reads, left-repeats, right-repeats and regular reads, respectively.

Precision-recall curve comparing the performance of both semi-supervised models is shown in Fig. 4.

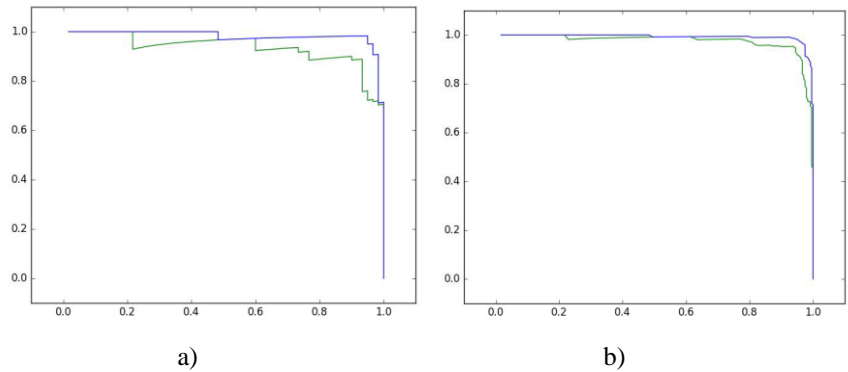

a) b)

**Fig. 4.** a) precision-recall curve for the chimeric class, b) mean precision-recall curve for all 4 classes. The blue graph represents the curve for semi-GAN, while the green one represents the curve for M1+M2. The x-axis represents recall, while the y-axis represent precision.

Finally, we extended the *de novo* genome assembly process by omitting the reads that were classified as chimeric and overlaps between the reads that were classified as left- and right-repeats. The overlap-layout-consensus approach [8] was used for genome assembly. We used the semi-GAN model as the classifier, and conducted an experiment over the following reference genomes: NCTC74 (Salmonella enterica subsp. enterica serotype Typhimurium), NCTC86 (Escherichia coli), NCTC129 (Salmonella enterica subsp. enterica serotype Newport), NCTC204 (Klebsiella pneumoniae sensu stricto). In Table 2. we compare the characteristics of the obtained overlap graphs generated by the extended process to the characteristics of the overlap graphs generated by the regular process which uses all available reads. In the comparison, we used two measures: the number of contigs and NG50 measure. Since it is very difficult to produce a contiguous sequence that covers the whole chromosome, assemblies are fragmented in uniquely determined sequence pieces. NG50 measure is another fragmentation measure that determines the size of a shortest contig which together with longer contigs covers at least 50 % of a reference genome.

**Table 2.** Characteristics of the obtained overlap graphs

| Genome | Extended process (no of contigs/NG50) | Regular process (no of contigs/NG50) |
|---|---|---|
| NCTC74 | 16 / 574k | 21 / 546k |
| NCTC86 | 53 / 187k | 126 / 74k |
| NCTC129 | 12 / 1008k | 36 / 270k |
| NCTC204 | 29 / 476k | 39 / 320k |

## 4   Discussion

Based on the results presented in Table 1., we can notice that both semi-supervised models achieved better results when compared with the model that had only used deficient labeled set. Therefore, we can draw a conclusion that unlabeled data in some way helps classifiers make better predictions.

We can also notice that semi-GAN outperforms M1+M2 model regardless of the number of labeled examples that was used during the training. However, this difference diminishes as we increase the number of labeled examples. We noticed that both models have problems detecting chimeric reads, so we compared them based on their precision-recall curves.

Fig. 4. shows that, based on the area under the curve, we can infer that the semi-GAN outperforms the M1+M2 model, regardless of the chosen threshold for detecting chimeric reads.

Considering the visualizations shown in Fig. 2. and Fig. 3., we can infer that the class groups associated with semi-GAN are more detached and coherent than the groups associated with the M1+M2 model. One of the reasons is the fact that semi-GAN uses extracted features as input in the classification part of the model, while the M1+M2 model's classification part is separated from the feature extraction, and is performed on the raw input.

Results in Table 2. show a difference between results obtained by using the extended and the regular process. Since the number of contigs can be understood as a measure of the complexity of the overlap graph, it can be observed that our model reduces that complexity, which can make the further steps of the genome assembly process simpler. Similarly, NG50 score can be viewed as a measure of assembly quality. We can see that our extended process achieves higher quality than the regular process. Considering all presented results, it can be inferred that the semi-GAN is the better classifier for this problem. Also, we show that machine learning approach for the detection of chimeric and repeat reads might reduce the number of contigs and thereby facilitate the assembly process.

## 5    Implementation details

Details about layers used in models are presented in the Table 3.

**Table 3.** Architectures that were used during the training of the described models with the chosen hyperparameters. conv K/N represents the convolution layer with kernel size K, and N filters, max-pool represents the max-pooling process, fc M represents the fully connected layer with M output features, bn represents the batch-normalization layer and conv-1 and max-pool-1 represent transpose layers that invert the convolution or pooling operation.

| FF | M1+M2 | semi-GAN |
|---|---|---|
| conv 5/16 | **m1-encoder** | **generator** |
| max-pool | conv 5/16 | fc 1600 |
| conv 3/32 | max-pool | bn |
| max-pool | conv 3/32 | conv 3/64$^{-1}$ |
| conv 3/64 | max-pool | bn |
| fc 256 | conv 3/64 | max-pool$^{-1}$ |
| fc 4 | fc 256 | bn |
| | fc 10 | conv 3/32$^{-1}$ |
| | **m1-decoder** | bn |
| | fc 8000 | max-pool$^{-1}$ |
| | conv 3/64$^{-1}$ | bn |
| | max-pool$^{-1}$ | conv 5/16$^{-1}$ |
| | conv 3/32$^{-1}$ | fc 100 |
| | max-pool$^{-1}$ | **discriminator** |
| | conv 5/16$^{-1}$ | conv 5/16 |
| | fc 500 | bn |
| | **m2-encoder** | max-pool |
| | fc 64 | bn |
| | fc 64 | conv 3/32 |
| | fc 3 | bn |
| | **m2-decoder** | max-pool |
| | fc 64 | bn |
| | fc 64 | fc 256 |
| | fc 10 | bn |
| | **m2-classification** | fc 1024 |
| | fc 64 | fc 4 |
| | fc 64 | |
| | fc 4 | |

## 6  Acknowledgments

We thank Robert Vaser for his constant assistance with handling the data, generating overlaps and running the assembly process.

This work has been supported in part by Croatian Science Foundation under the project UIP-11-2013-7353 "Algorithms for Genome Sequence Analysis".

## References

1. Baker, M.: De novo genome assembly: what every biologist should know. Nature Methods 9, 333-337 (2012).
2. Martin, J.A., Wang, Z.: Next-generation transcriptome assembly. Nature Reviews Genetics 12, 671-682 (2011)
3. Kingma, D.P., Rezende, D.J., Mohamed, S., Welling, M.: Semi-supervised learning with deep generative models. In: NIPS'14 Proceedings of the 27th International Conference on Neural Information Processing Systems, pp. 3581-3589. MIT Press Cambridge, Montreal, Canada (2014)
4. Goodfellow, I.J., Pouget-Abadie, J., Mirza, M., Xu, B., Warde-Farley, D., Ozair, S., Courville, A., Bengio, Y.: Generative Adversarial Networks. eprint arXiv:1406.2661 (2014)
5. Odena, A.: Semi-supervised learning with generative adversarial networks. eprint arXiv:1606.01583 (2016)
6. Sović, I., Šikić, M., Wilm, A., Fenlon, S.N., Chen, S., Nagarajan, N.: Fast and sensitive mapping of nanopore sequencing reads with GraphMap. Nature Communications 7, article number: 11307 (2016)
7. Van der Maaten, L.J.P., Hinton, G.E.: Visualizing High-Dimensional Data Using t-SNE. Journal of Machine Learning Research 9: 2579-2605 (2008)
8. Li, Z., Chen, Y., Mu, D., Yuan, J., Shi, Y., Zhang, H., Gan, J., Li, N., Hu, X., Liu, B., Yang, B., Fan, W.: Comparison of the two major classes of assembly algorithms: overlap-layout-consensus and de-bruijn-graph. Brief Funct Genomics 11 (1), 25-37 (2012)